\documentclass[conference]{IEEEtran}
\IEEEoverridecommandlockouts
\usepackage{cite}
\usepackage{amsmath,amssymb,amsfonts}
\usepackage{algorithmic}
\usepackage{graphicx}
\usepackage{textcomp}
\usepackage{xcolor}
\usepackage{comment}
\usepackage{multirow,multicol}
\def\BibTeX{{\rm B\kern-.05em{\sc i\kern-.025em b}\kern-.08em
    T\kern-.1667em\lower.7ex\hbox{E}\kern-.125emX}}
\begin{document}

\title{Adversarial Learning for Supervised and Semi-supervised Relation Extraction in Biomedical Literature\\
}

\author{\IEEEauthorblockN{ Peng Su}
\IEEEauthorblockA{\textit{Department of Computer and Information Science} \\
\textit{University of Delaware}\\
Newark, Delaware, USA \\
psu@udel.edu}
\and
\IEEEauthorblockN{K. Vijay-Shanker}
\IEEEauthorblockA{\textit{Department of Computer and Information Science} \\
\textit{University of Delaware}\\
Newark, Delaware, USA\\
vijay@udel.edu}

}

\maketitle

\begin{abstract}
Adversarial training is a technique of improving model performance by involving adversarial examples in the training process. In this paper, we investigate adversarial training with multiple adversarial examples to benefit the relation extraction task. We also apply adversarial training technique in semi-supervised scenarios to utilize unlabeled data. The evaluation results on protein-protein interaction and protein subcellular localization task illustrate adversarial training provides improvement on the supervised model, and is also effective on involving unlabeled data in the semi-supervised training case. In addition, our method achieves state-of-the-art performance on two benchmarking datasets.
\end{abstract}

\begin{IEEEkeywords}
Adversarial training, Semi-supervised learning, Relation extraction, Deep learning
\end{IEEEkeywords}

\section{Introduction}
With the explosion of text in the biomedical literature, a wealth of valuable knowledge hide in the text. Applying natural language processing (NLP) technique such as relation extraction can help knowledge base curation, which is an urgent problem given that manual curation will always lags behind the fast growth of literature. Deep learning models have great power to learn the representation of the training data and have made impressive success in many domains including natural language processing \cite{young2018recent}. Even though neural network models have achieved state-of-the-art performance on many problems, in some cases the performance is limited by the size of the training sets. Usually, the large amount of parameters in deep neural networks need large labeled dataset to train, but most new problems only have small labeled dataset available. The best way of acquiring large labeled dataset is to utilize human effort to put labels on the training data, but it is often not feasible since the labeling process could only be done by domain expertise. In this work, we will investigate to use adversarial learning to alleviate the problem of insufficient training data, specifically for relation extraction.

Adversarial method is a technique of designing malicious inputs to fool machine learning models \cite{kurakin2016adversarial}. Those malicious inputs (often called adversarial examples) are usually acquired by adding a small perturbation on the original inputs, and the process of model training using adversarial examples falls into the category of adversarial training (AT). Adversarial training in supervised learning scenario could strength the robustness and generality of the model since it involves both benign and malicious data in the training process. During the training, adversarial learning utilizes an extra loss on the adversarial examples using the same label as the corresponding example from the training data \cite{miyato2016adversarial}. 

A variation  of adversarial training called virtual adversarial training (VAT) was introduced in \cite{miyato2015distributional} to involve unlabeled data in the model training process. Other than adding adversarial examples to corresponding instances of the labeled dataset, adversarial examples are added on instances from a potentially larger unlabeled dataset. Because of the unlabeled aspect of these instances, an alternate loss function is defined. Both adversarial and virtual adversarial training can be seen as regularization method as an extra loss is added to the original loss of the model as a regularization term, which will be discussed in details later. Virtual adversarial training can also be seen as a semi-supervised learning method since it involves unlabeled data in the model training. 

Adversarial training technique was first introduced in the computer vision field \cite{goodfellow2014explaining}. In recent years, adversarial training has also been applied on tasks in NLP domain such as text classification \cite{miyato2016adversarial} and relation extraction \cite{wu2017adversarial}, but very limited work has been done. For example, in \cite{wu2017adversarial}, they only applied adversarial training on noisy data from MIML framework \cite{surdeanu2012multi} and only perturbation on specific part of the input features is explored. In addition, virtual adversarial training has been only applied on text classification task in NLP field, but not on relation extraction.   

In this paper, we apply adversarial training on relation extraction in the standard setting. As far as we know, this is the first work to introduce virtual adversarial training on the relation extraction tasks. To verify the effectiveness of our method, we will test it on two widely-studied relation extraction tasks in the BioNLP domain, the protein protein interaction (PPI) task \cite{krallinger2008overview} and the protein subcellular localization (PLOC) task \cite{kim2011overview} using three well-known benchmark datasets.

In summary, we investigate adversarial training in the standard setting of relation extraction as well as apply virtual adversarial training for relation extraction. We also conduct additional experiments that might shed light on adversarial and virtual adversarial training: a). involving multiple adversarial examples during training; b). adding perturbation on all input features of the model; c). exploring the size impact of unlabeled data. In addition, we note that we obtain leading results on two benchmark datasets after application of adversarial training.

\section{Related Work}
Relation extraction is typically seen as a classification task and current state-of-art systems on relation extraction are usually based on deep neural networks. Among all the deep learning models, two highly related network architectures are: convolutional neural networks (CNN) and recurrent neural networks (RNN). Both CNN models and RNN models have achieved notable results on relation extraction tasks \cite{nguyen2015relation,zeng2014relation,hsieh2017identifying,miwa2016end,su2019using}. Recently, many variants of them are also proposed to improve the performance by capturing more relation expression information. Piecewise-CNN (PCNN) \cite{zeng2015distant} applies piecewise max pooling process after the convolutional operation to extract more structural features between the entities. Multi-channel CNN model in \cite{peng2017deep} adds extra channel to capture the dependency information of the sentence syntactic structure, while the multi-channel CNN model in \cite{quan2016multichannel} integrates different versions of word embeddings to better represent the input words. Hua et al. \cite{hua2016shortest} build a deep learning model based on shortest dependency path (SDP), which is considered to contain the most important information of the relation expression. A residual CNN model is proposed in \cite{zhang2019exploring} and achieves comparable performance with other deep learning models on protein protein interaction task. In this work, we will experiment with PCNN model to illustrate the effectiveness of our method. 

Adversarial training is proposed by Goodfellow et al.  \cite{goodfellow2014explaining} to enable the model to classify both the original examples and adversarial examples on image classification task. They utilize adversarial examples to calculate an extra loss and add it on the original loss function to regularize the model. Before that, several machine learning methods, including deep neural networks, are found to be vulnerable to adversarial examples \cite{szegedy2013intriguing}, which are generated by adding small adversarial perturbation on the input. This vulnerability indicates that the input-output mappings learned by deep neural networks are fairly discontinuous, which means the model will misclassify the examples after adding a small perturbation. In the NLP domain, Xie et al.  \cite{xie2017data} derive a connection between input noising (random perturbation) in neural network language models and smoothing in n-gram models. Miyato et al.  \cite{miyato2015distributional,miyato2016adversarial} introduce adversarial training into text classification domain by applying the perturbation to the word embedding. Wu et al.  \cite{wu2017adversarial} apply adversarial training in relation extraction using distantly supervised data from multi-instance multi-label framework and they only perturb the word embedding part of the input to improve the robustness of the model. While this work creates adversarial examples by perturbing word embedding, we extend the work by investigating perturbing other features of the input as well. Additionally, we consider the impact of adding multiple adversarial examples at one time. Sometimes, adversarial training could be confused with the concept of generative adversarial networks (GAN) \cite{goodfellow2014generative}, which is a mechanism for training a new generative model to simulate the distribution of original data.

In order to involve unlabeled data, Miyato et al. \cite{miyato2015distributional,miyato2016adversarial} extend the adversarial training to virtual adversarial training by adding a local distributional smoothness regularization term on the model loss function. This method utilizes both the labeled and unlabeled data during training, so it could also be seen as a semi-supervised method \cite{zhu2009introduction}. Semi-supervised learning has recently become more popular since vast quantities of unlabeled data could be collected with low cost. The most common way to utilize semi-supervised learning is to acquire labels for unlabeled data and involve them in the model training process \cite{zhu2005semi}.The self-training scheme \cite{rosenberg2005semi} and the graph-based method \cite{chen2006relation,sterckx2016knowledge} belong to this kind of method. Recently, several novel methods have been proposed to involve unlabeled data for different tasks. Kingma et al. \cite{kingma2014semi} demonstrates that deep generative model and approximate Bayesian inference could provide improvement on image classification task. Graph neural network \cite{kipf2016semi} is also proven to be an effective semi-supervised method on classification problems. The virtual adversarial training method differs other semi-supervised by employing the adversarial training technique to help smooth the model and hence makes the model perform better \cite{miyato2015distributional}.  

\section{Methodology}

In this section, we start with the definition of relation extraction and a set of notations. Then we describe the architecture of the deep neural network model and the input representation of the model. Next, we discuss our proposal to use multiple adversarial examples during training. The application of virtual adversarial training is introduced at the end of this section.  

In this work, we detect the relation expression in a sentence. As is common, relation extraction is reduced to a binary classification problem given a sentence and the entity mentions. Hence, for a relation $R$, our model will predict the probability $P(R|e_1,e_2,w_1,\,w_2,\,w_3,\,\ldots,w_n)$ based on two entities $e_1$, $e_2$ within a sentence $S=w_1 \,w_2 \,w_3 \,\ldots w_n$ where $w_i$ represent the words in a sentence.

\subsection{PCNN Model Architecture}
As mentioned before, our investigations involve the use of piecewise CNN (PCNN) \cite{zeng2015distant}. Like regular CNN models for classification problem, it contains four different layers: a). convolution layer(s);  b). pooling layer(s); c). fully connected layer(s) and d). a softmax layer. The function of pooling layer(s) is to summarize the local features detected from the previous convolution layer(s) and then the summarized information will be used in fully connected layer(s) and softmax layer to classify each category. 

PCNN model differs from the standard CNN model in its pooling operation. The pooling operation of PCNN model is applied piece-wise based on the position of the entities in the sentence, thus including more structural information. Specifically, a sentence is divided into three parts using two entities as the segment points, and pooling is operated on these three parts separately. Let us take this sentence "We demonstrate that RB\textsubscript{PROTEIN} binds directly to hTAFII250\textsubscript{PROTEIN} in vitro and in vivo" as an example, we will do pooling on three parts: "We demonstrate that RB\textsubscript{PROTEIN}", "binds directly to hTAFII250\textsubscript{PROTEIN}", and "in vitro and in vivo". At last, we concatenate these three outputs obtained from the three separate pooling operations as the final output of pooling.

\begin{figure}[tb]
\centering
\includegraphics[width=0.45\textwidth]{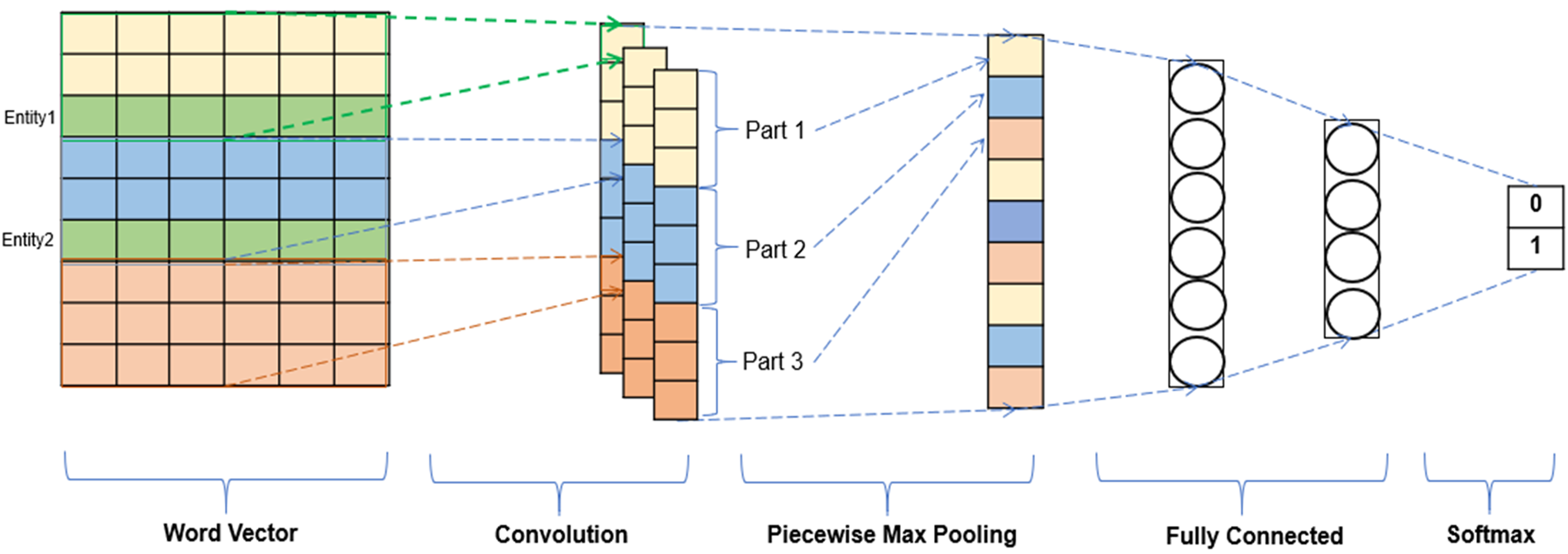}
\caption{\label{fig:pcnn} PCNN model architecture.}
\end{figure}

Fig. \ref{fig:pcnn} shows the structure of the piecewise CNN model and we use different colors to illustrate three parts of pooling operation.

\subsection{Word Representation}
As discussed below, for each word in a sentence in addition to the word embedding vector, we concatenate the POS tag, entity type, entity
positional information and incoming dependency relation information to form its vector representation.


Word embedding is usually learned on large corpus to represent the word better. Hence our choice in this paper is the pre-trained word embedding on the PubMed using skip-gram model \cite{chiu2016train} and the dimension of word embedding vector is 200. We extract POS tag and incoming dependency information from the parse results of Bllip parser \cite{mcclosky2010any} and covert them to unique 10-dimension vectors. For entity positional information, we calculate the relative distance to entities (to entity 1 and entity 2). Specifically, we count the words between the target word and the entities and the distance will be marked as negative distance if a word appears at the left side of the entity. At last, we will map each distance number to unique 5-dimension vector. As for entity type, all the words in the sentence could fall into four categories: Entity1, Entity2, Entity, O, where Entity1 and Entity2 are the two interacting entities, other entities in the sentence are marked as Entity, and O stands for other words. We use one-hot vector to represent this feature.

\subsection{Adversarial Training}

The idea behind adversarial training is that similar data instances should have same label, and deep neural network models should classify them in the same category. For each training instance, adversarial example is generated by adding a small perturbation on the original instance. Due to small perturbation, the new instance is seen as a similar instance and hence shares the label with the original instance. 

Formally, let $D_l$ be the manually labeled dataset used in training and $\theta$ is the parameters of the current model. If an instance $x$ in $D_l$ has a label $y$, then we will denote the generated adversarial example as $x+r_{adv}$, where $r_{adv}=arg\,\max \limits_{\|e\|\le \epsilon}\,L(x+e,y,\theta)$ and the hyperparameter $\epsilon$ bounds the magnitude of the perturbation. In regular training, the loss on the entire labeled dataset is computed: 
\begin{equation}
L(X,Y,\theta)=\sum_{(x,y)\in D_{l}}L(x,y,\theta)\label{eq1} 
\end{equation}
In adversarial training, the loss is computed as follows: 
\begin{align}
\tilde{L}(x,y,\theta)= L(x,y,\theta) + \alpha L(x+r_{adv},y,\theta)\label{eq2} \\
\tilde{L}(X,Y,\theta)=\sum_{(x,y)\in D_{l}}\tilde{L}(x,y,\theta)\label{eq3}\qquad    
\end{align}
 where $(X,Y)=D_l$ mean the set of inputs and labels respectively and $\alpha$ is usually set to 1.

\begin{figure}[tb]
\centering
\includegraphics[width=0.40\textwidth]{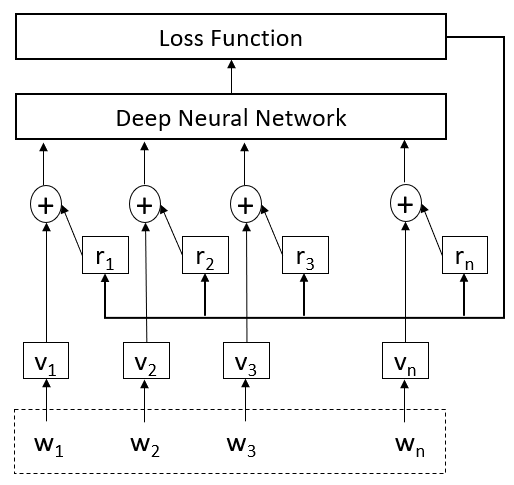}
\caption{\label{fig:arch}The model training with perturbation. $w_1,\,w_2,\,w_3,\,\ldots,w_n$ are the words in one sentence, $v_1,\,v_2,\,v_3,\,\ldots,v_n$ are the word vectors and $r_1,\,r_2,\,r_3,\,\ldots,r_n$ are the word vector perturbations which are updated through back-propagation at each training step.}
\end{figure}

At each training step, the adversarial perturbation $r_{adv}$ will be calculated first based on the current model setting, and then the perturbed example (adversarial example) will be feed into training as it is shown in Fig. \ref{fig:arch}. An extra term $L(x+r_{adv},y,\theta)$ is added on the original loss as is shown above, so adversarial training could also be seen as a regularization method from this perspective.

The optimization problem to calculate $r_{adv}$ (as defined above) is an intractable problem for deep neural networks. Goodfellow et al. \cite{goodfellow2014explaining} propose a linear (first-order Taylor) approximation method to calculate $r_{adv}$:
\begin{equation}
r_{adv}=\epsilon \frac{g}{\|g\|}\label{eq4}    
\end{equation}
where $g=\nabla_x L(x,y,\hat{\theta})$, which is easy to compute through backpropagation algorithm. We will use this approximation method in this paper.

In the adversarial training, the training data are increased by inclusion of the adversarial examples. Since the human-labeled dataset cannot be augmented, we can only increase training data by including more adversarial examples. Thus, we propose to add multiple adversarial examples for each instance in $D_l$. In this work, we will examine whether the addition of multiple adversarial examples can improve the generalization of model.

To generate extra adversarial examples, we just add even smaller random perturbations on the current adversarial example. Specifically, we generate a set of adversarial examples $D_{adv}$ by randomly generating perturbations $e_i$ and adding them on current adversarial example: $$D_{adv}=\{r_{adv}+e_i\}\quad i=0,1,...,M$$ where the magnitude of $e_i$ is much smaller than of $r_{adv}$, i.e., $\| e_i \| \ll \| r_{adv} \|$ and $e_0=0$ .

In the case of multiple adversarial examples, the loss function of adversarial training is:
\begin{align}
\tilde{L}(x,y,\theta)=L(x,y,\theta) + \alpha \sum_{r_i\in D_{adv}} L(x+r_{i},y,\theta)\label{eq5}\\ \tilde{L}(X,Y,\theta)=\sum_{(x,y)\in D_{l}} \tilde{L}(x,y,\theta) \label{eq6}\qquad\quad  
\end{align}
where $\alpha$ is set to 1 as before.

\subsection{Virtual Adversarial Training}

Semi-supervised learning is often applied when there is a small amount of labeled data but a large amount of raw (unlabeled) data are available. Note that in adversarial training, adversarial examples are generated only for instances in labeled dataset. Miyato et al. \cite{miyato2015distributional} propose semi-supervised inspired version of adversarial training called virtual adversarial training. We will now introduce this method in the relation extraction scenario.

Like before, let $x$ and $\theta$ represent the input of the model and the parameters of the model respectively. We are not assuming knowledge of label of $x$ ($x$ may be from the unlabeled set). As before, we will generate a perturbation $r_{vadv}$ for $x$. However, since the label of $x$ is not known, we use $p(x,\theta)$, the output distribution of predicted labels of the current model to compute the model loss. 

As both labeled and unlabeled data will be included in this method, let $D_l$ denote the labeled dataset and $D_{ul}$ represent the set of unlabeled data. The loss function $L_{vadv}$ is now defined as:
\begin{equation}
L_{vadv}(x,\theta)=KL(p(x,\theta)\| p(x+r_{vadv},\theta))\label{eq7}     
\end{equation}
where $r_{vadv}=arg\,\max \limits_{\|e\|\le \epsilon}\,L_{vadv}(x+e,\theta)$, $KL(p\|q)$ is the KL divergence of two distribution p and q, $\theta$ is the current parameter setting of the model and $\epsilon$ is the constraint of the perturbation magnitude. Hence, the loss function of virtual adversarial training is
\begin{equation}
\tilde{L}(X,Y,\theta) = \sum_{(x,y)\in D_{l}} L(x,y,\theta) + \lambda \sum_{x\in D_{ul}} L_{vadv}(x,\theta)\label{eq8}    
\end{equation}
where $\lambda$ is also usually set to 1. Virtual adversarial training examples can also of course be generated for instances from labeled set as well (by ignoring the label provided in the original dataset).

Next we discuss an approximation method to compute the virtual adversarial perturbation $r_{vadv}$ to overcome the intractability of solving the maximization problem of $r_{vadv}$. For simplicity, we denote $KL(p(x,\theta)\| p(x+r,\theta))$ by $K(x,r,\theta)$. We could utilize the same linear approximation as adversarial training before, however $L_{vadv}$ reaches its minimum value at $r_{vadv}=0$, which means $\nabla_r K(x,r,\theta)|_{r=0}=0$. Instead of using the first-order Taylor approximation, Miyato et al.  \cite{miyato2015distributional} suggest the use of second-order Taylor approximation: 
\begin{equation}
K(x,r,\theta) \approx \frac{1}{2}r^T H(x,\theta) r\label{eq9}    
\end{equation}
where $H(x,\theta)$ is the Hessian matrix of $K(x,r,\theta)$. Under this approximation: 
\begin{equation}
r_{vadv} \approx arg\,\max \limits_{\|r\|\le \epsilon}\,r^T H(x,\theta) r\label{eq10}    
\end{equation}
and $r_{vadv}$ will be the dominant eigenvector of $H(x,\theta)$ with magnitude $\epsilon$ since 
\begin{equation}
\max \limits_{\|r\|\le \epsilon}\,r^T H(x,\theta) r=\max \limits_{\|r\|\le \epsilon}\,r^T \lambda_d r=\lambda_d \epsilon\label{eq11}     
\end{equation}
where $\lambda_d$ is the dominant eigenvalue. However, it is computationally expensive to calculate the eigenvector and eigenvalue of a matrix.  Miyato et al. \cite{miyato2015distributional} utilize power iteration and finite difference method to approximate $r_{vadv}$. Let us denote $d$ is a random unit vector that is not perpendicular to the dominant eigenvector, and repeat the power method $I_p$ times as follows: 
\begin{equation}
d \leftarrow \overline{\nabla_r K(x,r,\theta)|_{r=\xi d}}\label{eq12}   
\end{equation}
where $\xi$ is a very small positive number and $\overline{\cdot}$ is normalization operation. Then $\epsilon d$ will be good approximation of dominant eigenvector of $H(x,\theta)$, i.e. $r_{vadv}$.

\section{Experiments}
In this section, we will design experiments to evaluate the adversarial and virtual adversarial training method using human-labeled datasets from two tasks.

As discussed previously, we will introduce multiple adversarial examples in the adversarial training. Thus, the first set of experiments focus on adversarial training and the effect of adversarial example number. Specifically, we will see how the models perform by adding different number of adversarial examples in the training process. As more features are added as the input of the deep neural network, we also plan to explore the effect of adversarial perturbation not only on word embedding but also on extra features of the input.

Our next set of experiments will apply virtual adversarial training on relation extraction task. Since there is no theory to guide us how much unlabeled data we should use in the virtual adversarial training, we will also considers the size effect of unlabeled dataset in the virtual adversarial training setting. Specifically, we will evaluate the models built with different size of unlabeled data and find the appropriate setting in virtual adversarial training to guide the generation of unlabeled data.  

Although we propose to use virtual adversarial training on relation extraction, we have not discussed anything about the unlabeled data. Specifically, we are interested in knowing whether the similarity of the unlabeled dataset to the labeled set might impact the results. While it is a hard question to know how similar these two datasets are, we investigate a simple situation where the two sets are the same or there are no examples in common. In the former case, $D_{ul}$ will contain the same elements as the $D_l$ except that in $D_{ul}$, the instances are seen as unlabeled (by ignoring the labels). In discussing the results, we will use VAT* for this case and utilize VAT when $D_{ul}$ is different with $D_l$.

\subsection{Labeled Datasets}

AIMed \cite{bunescu2005comparative} and BioInfer \cite{pyysalo2007bioinfer} are two widely used benchmark dataset for PPI task, and LocText \cite{cejuela2018loctext} is a recently available human-labeled corpus for PLOC task. We will use them as our labeled datasets, and Table~\ref{tab:statistics} shows the statistics of these three datasets.

\begin{table}[tb]
\caption{Statistics of AIMed, BioInfer, LocText and unlabeled data.}
\centering
\begin{tabular}{ c c c c }
\hline

\hline

\hline

\hline
Dataset &  Positive\# & Negative\# & Unlabeled\#\\
\hline
AIMed &  1,000 & 4,834 & 136,687\\
BioInfer & 2,534 & 7,132 & 136,687\\
LocText &  351 &338 & 111,120\\
\hline

\hline

\hline

\hline
\end{tabular}
\label{tab:statistics}
\end{table}

\subsection{Generation of Unlabeled Data}

In order to generate unlabeled examples for our relation extraction tasks, we have to find a text source and a method to recognize all entity names in the text. The literature found in the IntAct database \cite{orchard2013mintact} is large enough as our text source for PPI. For entity names, we utilize the end-to-end system GNormPlus \cite{wei2015gnormplus} to detect gene/protein names. For the subcellular location names, we use location names from UniProt \cite{uniprot2018uniprot} as a dictionary to match the mentions in the Medline text, which is the text source for PLOC task. Once we have all the entities recognized in the text, we could generate the unlabeled data by considering every possible combination of entity names in one sentence. In this way, we generate a large amount of unlabeled examples for PPI and PLOC (see Table \ref{tab:statistics}), but we only use part of those examples in our experiments.

\subsection{Experimental Setup}

All our experiments are implemented in Tensorflow \cite{abadi2016tensorflow} and the input sentence length is set to 100 (longer sentences are pruned and the shorter sentences are padded with zeros). For the PCNN model, we utilize filter number of 400 and convolution window size of 3. Training epoch is set to 200 in all the experiments. In the adversarial training experiments, we use learning rate of 0.001 with 0.95 decay rate and 1000 decay steps, batch size of 128 and adversarial perturbation $\epsilon = 0.01$. For the semi-supervised learning experiment, learning rate of 0.001, batch size of 128 for labeled data and batch size of 128, 256 and 384 for unlabeled data are used in the training. For the adversarial perturbation $\epsilon$ in semi-supervised learning, we use the value of 2 for word embedding and 0.01 for the rest features. Since the purpose of this perturbation is used to promote the smoothness of the model, bigger value for word embedding is used than the one in adversarial training.

The magnitude (or norm) of word embedding is another thing we have to consider because it will affect the choice of perturbation magnitude. In order to reduce the influence of embedding norm, we normalize the pre-trained word embedding vector $v_i$ in adversarial training using a simple trick: 
\begin{equation}
v_i=\frac{v_i}{N_{max}}\label{eq13}    
\end{equation}
where $N_{max}$ is the maximum value in all pre-trained embedding vectors. In this way, we restrict the embedding norm to a relatively small range.

\section{Results and Discussion}
  
\begin{table*}[tb]
\caption{PCNN model experiment results on PPI and PLOC task. }
\centering
\begin{tabular}{ c c c c | c c c | c c c c } 
\hline

\hline

\hline

\hline
\multirow{3}{4em}{Method} & \multicolumn{6}{c|}{PPI} & \multicolumn{3}{c}{PLOC}\\\cline{2-10}

& \multicolumn{3}{c|}{AIMed} &\multicolumn{3}{c|}{BioInfer} & \multicolumn{3}{c}{LocText}\\\cline{2-10}
& Precision & Recall&F score& Precision & Recall&F score& Precision & Recall&F score \\ 
\hline
PCNN &75.6  &  75.6 & 75.6 &84.4&88.5&86.4& 73.6  & 84.7& 78.8 \\
\hline
PCNN+ADV\textsubscript{we}& 77.3 &  76.5 & 76.9 &85.4&88.8&87.1&  75.6 & 83.8 & 79.5\\
\hline
PCNN+ADV\textsubscript{Multi}& 77.8 &  \textbf{77.0} & \textbf{77.4}&85.6&\textbf{89.1}&\textbf{87.3} &  \textbf{75.9} & 84.1 & \textbf{79.8}\\
\hline
PCNN+VAT&  \textbf{78.1} & 74.1 & 76.1 &\textbf{88.3}&80.3&84.1& 74.0& \textbf{85.3} & 79.3\\
PCNN+VAT*&  77.4 & 75.0 & 76.2 &85.1&88.6&86.8& 75.0& 83.8 & 79.1\\

\hline

\hline

\hline

\hline
\multicolumn{10}{p{13.5cm}}{PCNN is the model when we add POS tag, Entity type and dependency information besides the word embedding and positional information in the input; PCNN+ADV\textsubscript{we} is the model when adversarial perturbation is only added on word embedding; PCNN+ADV\textsubscript{Multi} is the model when we use two adversarial examples in the training; PCNN+VAT* is the model when we involve unlabeled data from only labeled data (ignore the label) in the training; PCNN+VAT is the model when we involve unlabeled data (same size with labeled data) from $D_{ul}$ in the training.}
\end{tabular}
\label{tab:results}
\end{table*}

We use precision, recall and F score as the measurement of model performance and 10-fold cross validation is performed in these experiments. In order to reduce the effect of random initialization of weights in the training of deep neural network, we run 3 times for each experiment and take the average of these 3 rounds as the final results.

\subsection{Adversarial Training}
Our first experiment explores the effect of adversarial training where the input is perturbed. The first row of Table \ref{tab:results} provide the results for our baseline model, i.e., with the basic PCNN with no perturbation to the input. Since the text input is decided by the words and their order, we first perturb only the word embedding part of the input. Row 'PCNN+ADV\textsubscript{we}' of Table \ref{tab:results} gives the results showing improvement over the baseline for all three datasets.

\subsection{One vs Multiple Adversarial Examples}
As we discussed before, adding multiple adversarial examples might help the model generalize better since this training technique covers more nearby points of the original training example. Also, the current adversarial example is calculated by an approximation method, involving more examples might compensate for the approximation loss. 

\begin{figure}[tb]
\centering
\includegraphics[width=0.45\textwidth]{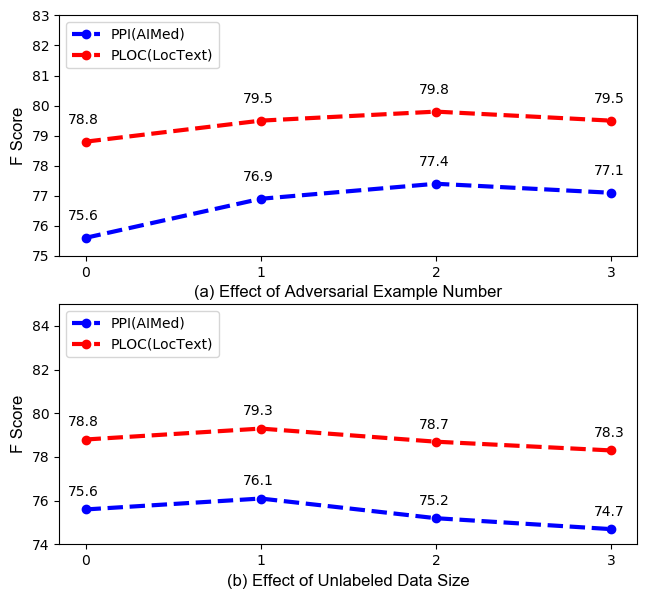}
\caption{\label{fig:advn}(a). The effect of adversarial example number; It shows the model performance with different number of adversarial examples. (b). The effect of unlabeled data size; It illustrates the model performance with different size of unlabeled data, where the X axis is the size ratio between unlabeled data and labeled data, i.e. $\frac{|D_{ul}|}{|D_l|}$.}
\end{figure}

In our experiments, we add two or three (i.e. $M=2,3$, where M is the size of $D_{adv}$ from previous section) adversarial examples for each instance of $D_l$. Please note that rows 'PCNN' and 'PCNN+ADV\textsubscript{we}' in Table \ref{tab:results} cover the case of $M=0$ and 1 respectively. Row 'PCNN+ADV\textsubscript{multi}' of the Table \ref{tab:results} shows the results for $M =2$ where we obtain the best results and this finding is the same for all three datasets. The results for $M=3$ are slightly worse than the results for $M =2$ and about the same as $M=1$ (see Fig. \ref{fig:advn}a). We believe this indicates that an appropriate number of  adversarial  examples  could  help  improve the  performance, but that increasing the number of adversarial examples further will introduce extra noise, negatively affecting the performance. More investigation is needed to shed light on how much to increase the number of adversarial example.

\subsection{Perturbation of Input}
As is discussed previously, we first perturbed only the word embedding to modify the text input. Notice that the remaining features are all inferred from these words by taggers, named-entity recognition (NER) tools and parsers. However, it is natural to consider perturbing not only the word embedding but  also the other features of the input. Compare to the performance of perturbing only the word embedding (row 2 of Table \ref{tab:results}), the results drop for the perturbation of all features: F score of 76.7, 87.0 and 79.1 for AIMed, BioInfer and LocText respectively. These numbers are slightly lower than the corresponding number of 76.9, 87.1, and 79.5 (row 2 of Table \ref{tab:results}) respectively. We speculate that since the length of vector to represent the other features are small, the  perturbation  on these features might significantly affect the relation representation in the input vector. This finding suggests that we need to be careful in deciding how the input should be perturbed when dealing with text.

\begin{table*}[tb]
\caption{Performance comparison with previous models.}
\centering
\begin{tabular}{c c c c | c c c c}
\hline

\hline

\hline

\hline
\multirow{2}{4em}{Model}&\multicolumn{3}{c|}{AIMed}&\multicolumn{3}{c}{BioInfer}\\
\cline{2-7}
&  Precision & Recall & F score& Precision & Recall & F score \\
\hline
PCNN\textsubscript{raw} \cite{zeng2015distant} &75.0  &  75.7 & 75.3 &83.9&88.1&85.9 \\
sdpCNN \cite{hua2016shortest} & 64.8& 67.8& 66.0& 73.4& 77.0& 75.2\\
MCCNN \cite{quan2016multichannel}&76.4&69.0&72.4&81.3&78.1&79.6\\
TK+WE \cite{li2015approach} & - & - & 69.7& - & - & 74.0\\
DSTK \cite{murugesan2017distributed} & 68.9 & 73.2 & 70.0 & 75.7 & 76.9 & 76.3\\
BiLSTM \cite{hsieh2017identifying} &  \textbf{78.8} & 75.2 & 76.9&  \textbf{87.0} & 87.4 & 87.2\\
BiLSTM\textsubscript{our}  &  75.6 & 70.2 & 72.8&  83.4 & 84.5 & 83.9\\
\hline
Ours &77.8 &  \textbf{77.0} & \textbf{77.4} &85.6 &  \textbf{89.1} & \textbf{87.3}\\
\hline

\hline

\hline

\hline

\multicolumn{7}{p{9.5cm}}{PCNN\textsubscript{raw} is the original model in \cite{zeng2015distant} whose input only contains word embedding and entity positional information. BiLSTM\textsubscript{our} stands for our evaluation results on BiLSTM model in \cite{hsieh2017identifying} using our parsed data on standard dataset. The results reported in \cite{hsieh2017identifying} are based on dataset where the authors did not use the standard number of instances.} 
\end{tabular}
\label{tab:comp}
\end{table*}
\subsection{Virtual Adversarial Training}
In the virtual adversarial training setting, we utilize both the labeled and unlabeled data to train the model. The addition of unlabeled data provides possible inputs for us to better smooth the model. The results are shown in the penultimate row ('PCNN+VAT') of Table \ref{tab:results}. 

Although we improve on the baseline model for AIMed and LocText, it is surprising that on BioInfer there is significant drop. We wonder whether this is due to a significant difference in the textual and distributional characteristics of the labeled and unlabeled text data. For this reason, we conducted another experiment where we wish to avoid this case. The simplest way to do this is to use the labeled data as the text for the unlabeled datasets, i.e. use the same text but drop the label. The results are shown in the last row ('PCNN+VAT*') of Table \ref{tab:results}, and are slightly better than before. In particular, there is a significant improvement for BioInfer where once again we improve on the baseline.

In the virtual adversarial training method, there is no prescribed choice for the size of unlabeled data relative to the labeled data. In our experiment (previously described results), we used the same size for the two dataset, i.e. $|D_{ul}|=n|D_l|$ where $n=1$. To investigate the possible impact of changing unlabeled data size, we conduct two more experiments with $n=2$ and $n=3$. As is shown in Fig. \ref{fig:advn}b, it appears that the choice of $n=1$ is the best and that bigger unlabeled data size takes away from the training to learn labels, negatively affecting the model performance.
\subsection{Comparison with Other Systems}
We have shown that adversarial and virtual adversarial training method improve the performance of PCNN model. We wish to place these results in a broader context of the results on these benchmark sets reported in the literature. To the best of our knowledge, there is no machine learning model trained on the LocText dataset for PLOC task except the work \cite{su2019using}, but they utilize external knowledge base by applying transfer learning. Thus, Table \ref{tab:comp} only compare our method with previous machine learning models on PPI task. 

As it is shown in Table \ref{tab:comp}, our system utilizing PCNN model and adversarial training technique achieves the state-of-the-art performance on the standard AIMed and BioInfer datasets of PPI task.

In addition, we are aware that there are other deep learning models built for PPI task, but our methods are not comparable to them due to three reasons. a). different evaluation metric; DCNN model in \cite{choi2018extraction}, BiLSTM model in \cite{yadav2019feature} and treeLSTM model in \cite{ahmed2019identifying} employ macro F score to evaluate their model, which is usually used on multi-class classification problem. Furthermore, the unbalanced evaluation corpus (Positive:Negative=1:4.8 in AIMed and Positive:Negative=1:2.8 in BioInfer) will make the macro F score much higher than normal F score. b). non-standard evaluation set; For example, in\cite{zhang2019deep}, the authors delete nested entities interaction from original corpora. c). different cross validation method; The paper of McDepCNN model \cite{peng2017deep} utilizes document-level during evaluation, while the models in Table \ref{tab:comp} use instance-level evaluation \footnote{Document-level evaluation means the instances from the same document could appear in the training set or test set only (not appear these two sets at the same time). Instance-level evaluation does not have this restriction during the cross validation evaluation process.}.

\subsection{Applying AT/VAT with RNN}
As shown in Table \ref{tab:results} that adversarial training is an effective way to boost PCNN model performance. In order to verify that our method generalize beyond CNN-based model, we also test our method on an RNN-based model. In particular, we repeat all our investigations on the BiLSTM model from \cite{hsieh2017identifying}, using the same input representation. We observe that adversarial and virtual adversarial training also improve the BiLSTM model performance. However, the BiLSTM model provides a better baseline, it does not  achieve state-of-the-art performance. Due to space reasons, we do not provide the details for the BiLSTM model.
\section{Conclusion}
In this paper, we utilize adversarial training to alleviate the problem of insufficient training data of deep learning models and promote the model generalization. We apply this technique on relation extraction tasks and extend it with multiple adversarial examples in the training process. The experiment results illustrate that adversarial training could improve the model performance and one more extra adversarial example could further boost the performance.

We also apply adversarial training technique in semi-supervised learning (virtual adversarial training) to utilize unlabeled data that acquired with low cost. The performance shows improvement when only small amount of unlabeled data are used in the semi-supervised training, but drops when the large volume of unlabeled data are involved. In addition, the performance of virtual adversarial training method is not up to par with adversarial training method in supervised learning scenario. 

In the future, we plan to explore the unlabeled data size effect on a larger scale to better guide its use. Also, We will continue to pursue better adversarial example generation technique to acquire more useful examples during the training to help the model generalization.

\section{Acknowledgement}
This work is funded by the grant U01GM120953 from National Institute of General Medical Sciences. The funders had no role in study design, data collection and analysis, decision to publish, or preparation of the manuscript.

\bibliographystyle{./bibliography/IEEEtran}
\bibliography{./bibliography/IEEEabrv,./bibliography/IEEEexample}

\end{document}